\documentclass[10pt,twocolumn,letterpaper]{article}

\usepackage{cvpr}              

\usepackage[dvipsnames]{xcolor}
\definecolor{cvprblue}{rgb}{0.21,0.49,0.74}
\usepackage[pagebackref,breaklinks,colorlinks,citecolor=cvprblue]{hyperref}
\usepackage{xcolor}
\usepackage{multirow}
\usepackage{makecell}
\def\paperID{315} 
\def\paperID{315} 
\def\confName{3DV\xspace}
\def\confYear{2024\xspace}

\title{3DRef: 3D Dataset and Benchmark for\\ Reflection Detection in RGB and Lidar Data}

\author{Xiting Zhao \qquad Sören Schwertfeger\\
ShanghaiTech University, Key Laboratory of Intelligent Perception
and Human-Machine Collaboration –\\ ShanghaiTech University, Ministry of
Education, China\\
{\tt\small \{zhaoxt, soerensch\}@shanghaitech.edu.cn}
}

\begin{document}
\includegraphics[page=1,scale=0.8]{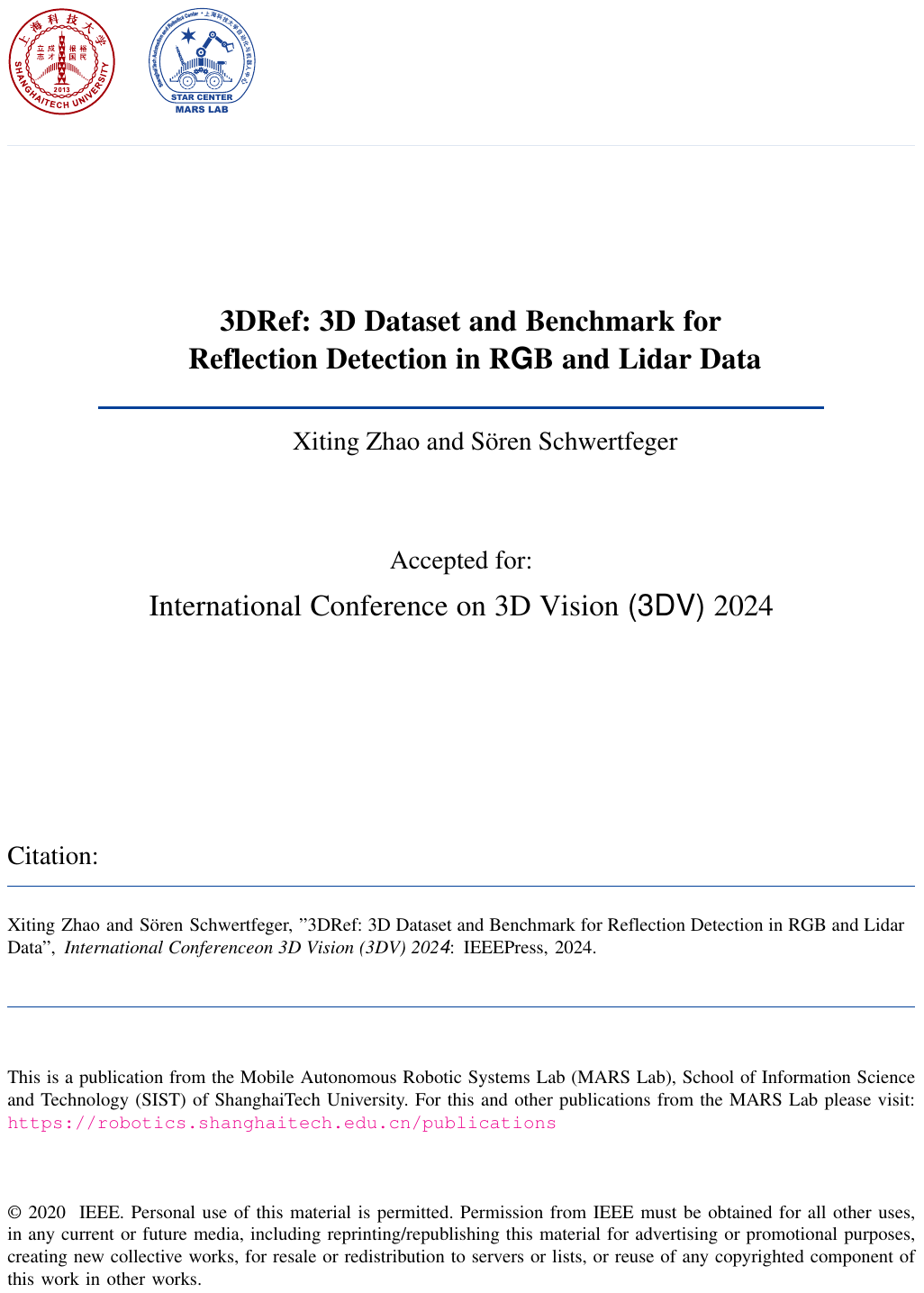}
\maketitle
\begin{abstract}
Reflective surfaces present a persistent challenge for reliable 3D mapping and perception in robotics and autonomous systems. However, existing reflection datasets and benchmarks remain limited to sparse 2D data. This paper introduces the first large-scale 3D reflection detection dataset containing more than 50,000 aligned samples of multi-return Lidar, RGB images, and 2D/3D semantic labels across diverse indoor environments with various reflections. Textured 3D ground truth meshes enable automatic point cloud labeling to provide precise ground truth annotations. Detailed benchmarks evaluate three Lidar point cloud segmentation methods, as well as current state-of-the-art image segmentation networks for glass and mirror detection. The proposed dataset advances reflection detection by providing a comprehensive testbed with precise global alignment, multi-modal data, and diverse reflective objects and materials. It will drive future research towards reliable reflection detection. The dataset is publicly available at \url{http://3dref.github.io} 
\end{abstract}   

\section{Introduction}
\label{sec:introduction}
Detecting reflective surfaces such as glass and mirrors presents a long-standing challenge in 3D computer vision and robotics. Frequently, for mobile robots employing 2D Lidar for mapping and navigation, spurious reflections from glass or mirrors give rise to "phantom" walls in occupancy grid maps. This phenomenon leads to motion planning failures, as the robot tends to avoid traversing these false obstacles \cite{tibebu2021Lidar}. Similarly, robot systems relying on depth or RGB cameras also encounter difficulties when scanning reflective objects like glass or shiny surfaces.

While applications like autonomous vehicles primarily operate in outdoor environments with less pervasive reflective surfaces, indoor settings like offices, homes, and factories introduce a higher degree of reflective challenges. The reflections from water, mirrors, glass, and windows remain problematic, particularly for applications such as inventory robots, service robots, and autonomous delivery vehicles, which require accurate mapping and localization to function reliably in reflections environment. For instance, glass buildings tend to generate phantom walls, thereby constraining drivable space estimates. Therefore, to perform their tasks reliably, all robotic systems across industrial, urban, and domestic settings must effectively handle reflective surfaces.
\begin{figure}[ht]
\centering
\includegraphics[width=0.4\textwidth]{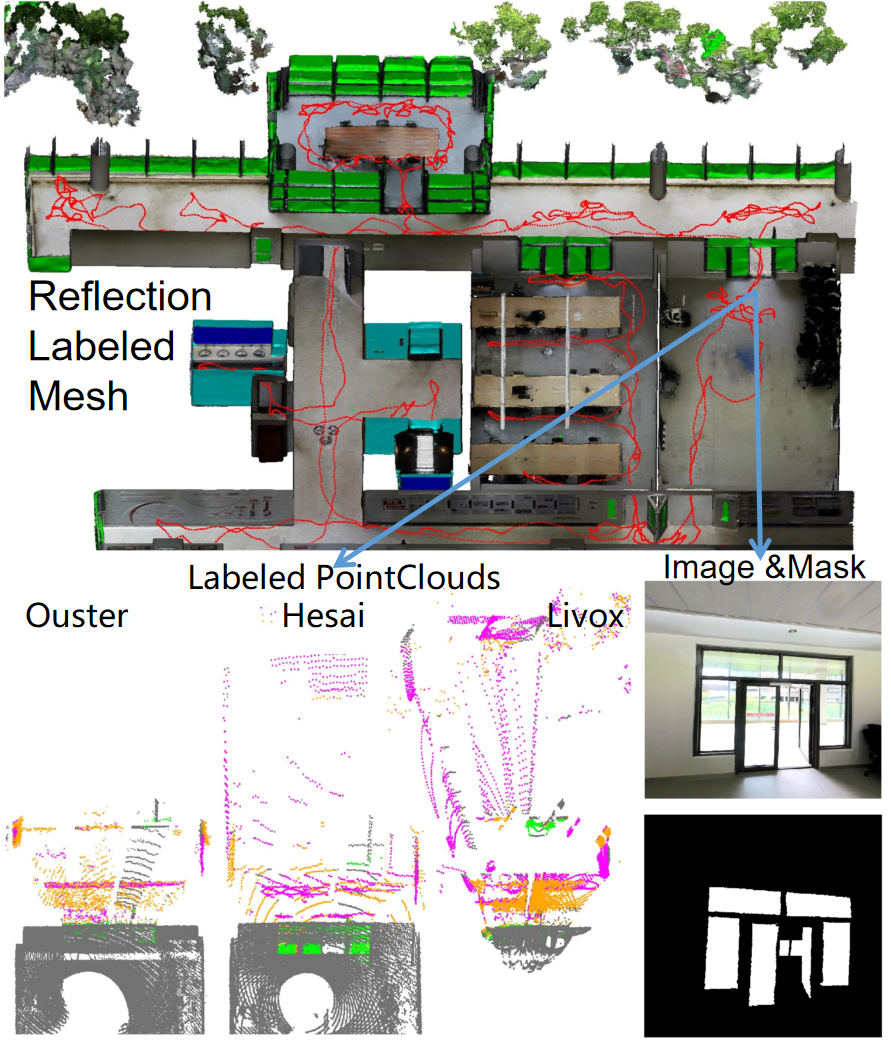}
\caption{The 3DRef dataset uses a labeled mesh to provide multi-modal reflection detection data, including three different multi-return Lidar and RGB image with mask. The label across all reflective material including glass, mirror and other reflective objects.}
\label{fig:dataset}
\end{figure}

Recent trends indicate increasing utilization of 3D Lidar sensors such as Ouster and Livox for 3D mapping and odometry \cite{jung2023asynchronous}\cite{liosam2020shan}. The advent of 3D Lidars like Velodyne, Ouster, and Livox unlocked the capability to sense a scene from various view points, potentially enabling reflection disambiguation. However, mirror-like reflections remain problematic for 3D Lidar sensors. Reflective objects defy the prevalent assumption in Lidar processing, which argues that each laser pulse generates a return point by reflecting off the closest surface in its path. Conversely, mirrors can cause secondary reflections when the beam continues traveling and reflects off other objects. Factors such as distance, incident angle, Lidar return strategy, and material property significantly influence the Lidar point result. Lidar sensors may return the point on reflective material, the obstacle behind transparent reflective material, the wrong reflection point, or fail to return any point due to the laser being absorbed by the material \cite{zhao2020mapping}, leading to phantom points that severely impair sensor models used for mapping, localization, and navigation.

The surge towards autonomous robots at scale has revived interest in detecting reflections. Detecting and handling reflective surfaces has been an active area of research in robotics for decades. Several approaches using Lidar, RGB camera, RGBD camera, and sensor fusion methods, have been proposed to address this challenge. However, these mostly operate in 2D, including 2D images, 2D maps, and 2D Lidar \cite{mo2023survey}.
Modern multi-return 3D Lidars provide new opportunities, capturing multiple returns per laser pulse reflecting off different object along its path.  However, most current methods only utilize one return, missing critical information in intermediate bounces. The return numbers of a point in a Lidar point cloud indicate which return the pulse belongs to. This information can help identify different kinds of objects and surfaces, such as glass and reflections, that might otherwise be confused.  
Our work aims to advance robust reflection detection by introducing the novel 3D Lidar multi-modal reflection detection dataset and a benchmark survey of the state-of-the-art techniques.

Our main contributions are as follows:
\begin{itemize}
\item A novel diverse benchmark dataset containing 51800+ aligned data samples across RGB, Lidar, with labels for 3D reflection detection in indoor environments.
\item Standardized ground truth representation via textured 3D meshes and automatic point cloud labeling to precisely annotate different reflective surface types beyond 2D masks.
\item Benchmarking state-of-the-art reflection detection methods to evaluate Lidar and RGB approaches, analyzing factors like multi-return pulses.
\end{itemize}

The paper is structured as follows. Section \ref{sec:introduction} provides an introduction to the challenges of reflection detection and motivation for a multi-modal 3D dataset. Section \ref{sec:related} reviews related work on existing reflection datasets and detection methods. Section \ref{sec:data} details our data collection platform and process. Section \ref{sec:dataset} describes our proposed dataset including the annotation methodology and statistics. Section \ref{sec:benchmark} presents benchmark results assessing current Lidar and RGB-based approaches on our new benchmark. Finally, Section \ref{sec:conclusion} concludes with a summary and directions for future work.
\section{Related Works}
\label{sec:related}
We consider prior work in two related topic areas: reflection datasets and reflection detection methods.
\subsection{Reflection Datasets}
\begin{table}[h]
\centering
\resizebox{\linewidth}{!}{
\begin{tabular}{cccccc}
 \hline
Dataset & Modalities & Samples & Objects \\ \hline
GDD\cite{mei2020don} & RGB & 3900 & glass \\
GSD\cite{lin2021rich} & RGB & 4102 & glass \\
RGBP-G\cite{mei2022glass} & RGB-P & 4511 & glass \\  \hline
MSD\cite{yang2019my} & RGB & 4018 & mirror \\
PMD\cite{lin2020progressive} & RGB & 6461 & mirror \\
RGBD-M\cite{wang2018depth} & RGB-D & 3049 & mirror \\
Mirror3D\cite{tan2021mirror3d} & RGB-D & 5894 & mirror \\  \hline
TROSD\cite{sun2023trosd} & RGB-D & 11060 & mirror\&glass \\ \hline
3DRef (Ours) & Lidar, RGB & 48024, 3799 & all reflective\\
 \hline

\end{tabular}
}
\caption{Comparison with existing glass and mirror object datasets}
\label{tab:comparedataset}
\end{table}
Although there are many Lidar indoor and outdoor datasets \cite{geiger2013vision,ramezani2020newer,gao2022vector}, none of them contain labels for reflective surfaces. For RGB images, several datasets have been introduced for glass and mirror detection using deep learning, as summarized in Table \ref{tab:comparedataset}. However, most only contain one kind of 2D image masks rather than full 3D annotations. Some use limited samples from existing RGB-D datasets like Sun RGB-D \cite{song2015sun} and Scannet \cite{dai2017scannet}. Our proposed dataset significantly advances the scope compared to prior work by providing a large-scale aligned 3D dataset with over 51800 samples across RGB images, multi-return Lidar point clouds, and semantic labels for different reflective surface types.
\subsection{Reflection Detection Methods}
\subsubsection{Reflection Detection Methods Using 3D Lidar}
Several works have explored the detection and removal of reflective surfaces from 3D Lidar point clouds. Gao et al. \cite{gao2021reflective} proposed filtering reflective noise from large-scale 3D point clouds collected from multiple Lidar positions. Yun et al. \cite{yun2018reflection,yun2019virtual} introduced methods to identify and remove reflective points from large-scale 3D point clouds. But these three papers all deal with Terrestrial Laser Scanners and not the widely used 3D Lidar sensors. Koch et al. \cite{koch2017detection} detected and removed specular reflections in 3D range measurements by analyzing triple return pulses from a rotating 2D Lidar. More recently, Henley et al. \cite{henley2023detection} used multi-bounce returns from a specialized multi spot Lidar system to detect and map specular surfaces in 3D. Zhao et al. \cite{zhao2020mapping} demonstrated detecting and utilizing Lidar reflections for mapping using a Velodyne 3D Lidar utilizing each dual return point cloud and using plane fitting and the intensity heap method to detect glass planes. These three papers demonstrate that leveraging the multi-return capability of Lidar sensors can benefit reflection detection. Foster et al. \cite{foster2023reflectance} propose the Reflectance Field Map, based on the concept of neural light fields from computer graphics. The method does not rely on intensity or multi-return measurements can handle dynamic environments and different surfaces of reflectivity and transparency. But it still use 2D map for evaluation.

\subsubsection{Reflection Detection Methods Using RGB Cameras}
A lot of learning based methods have been proposed for glass and mirror detection in images. Lin et al. \cite{lin2022exploiting} exploit semantic relations for glass surface detection. Earlier Lin et al. \cite{lin2021rich} proposed aggregating rich context with reflection priors for glass detection. Hu et al. \cite{hu2023tgsnet} use multi-field feature fusion and Transformers for glass segmentation. He et al. \cite{he2023efficient} develop multi-level heterogeneous learning for efficient mirror detection. Guan et al. \cite{guan2022learning} learn semantic associations to detect mirrors, while Huang et al. \cite{huang2023symmetry} employ symmetry-aware Transformers to detect mirrors.

\subsubsection{Other Reflection Detection Methods}
Additionally, several other approaches employ sensor fusion or integrate multiple modalities like depth, or polarization to detect reflections. Lin et al. \cite{lin2022depth} proposed depth-aware glass surface detection with cross-modal context mining using RGBD data. Mei et al. \cite{mei2022glass} present glass segmentation using intensity and spectral polarization cues from polarization cameras. Tao et al. \cite{tao2023glass} develop a glass recognition and map optimization method for mobile robots based on boundary guidance with sensor fusion. Han and Sim \cite{han2022zero} explore zero-shot learning for reflection removal in 360-degree images. Wei et al. \cite{wei2018multi} fusion 2D Lidar with ultrasound to detect the glass.

\section{Data Collection Platform}
\label{sec:data}

To collect the reflection dataset, we employed a diverse range of sensors listed in Table \ref{tab:sensors}, including Ouster OS0-128 Lidar, Livox Avia triple return Lidar, and Hesai Pandar qt64 dual return Lidar. Additionally, we used an Insta360 camera to capture RGB images of the surroundings.
\begin{figure}[htbp]
\centering
\includegraphics[width=0.4\textwidth]{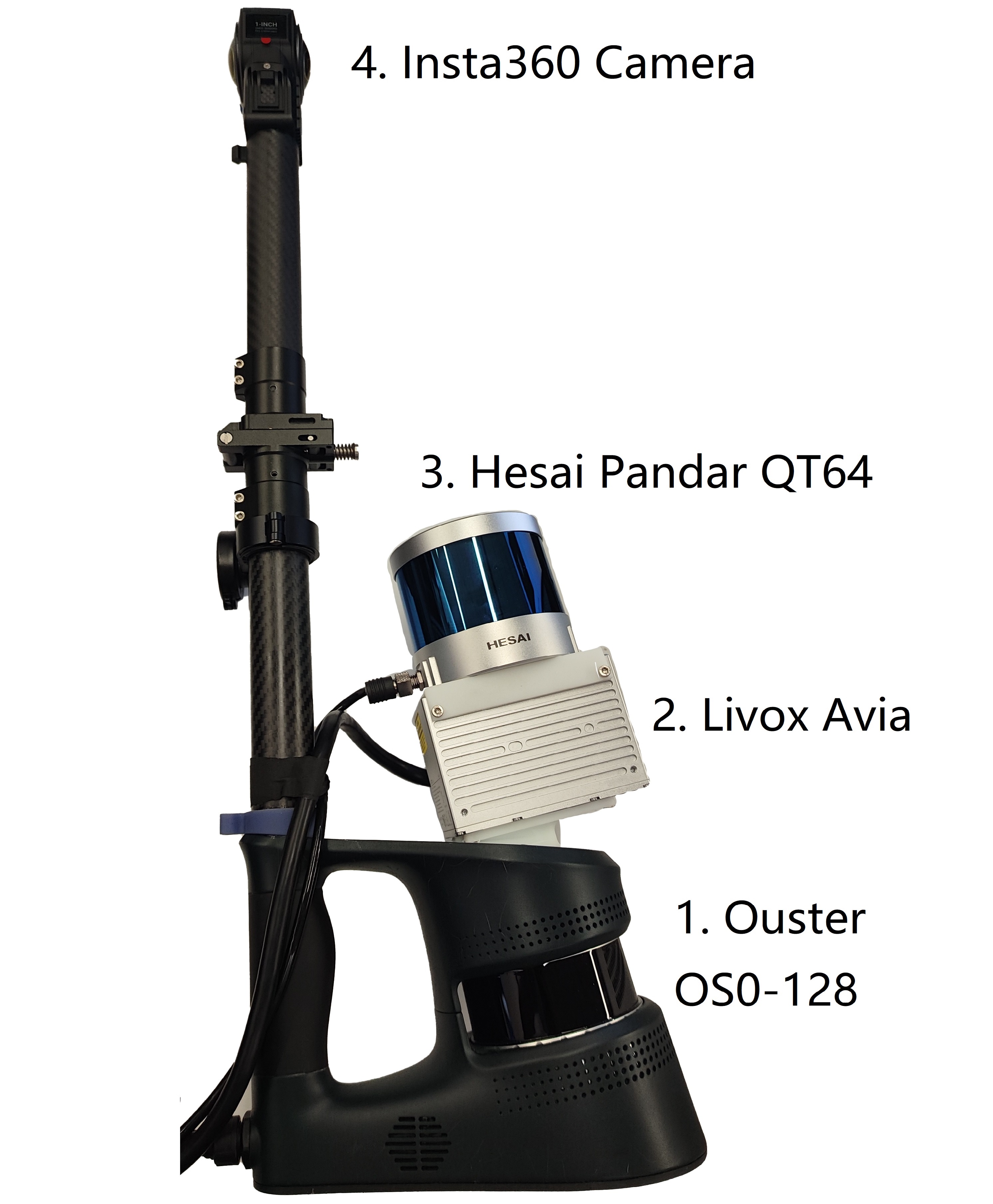}
\caption{Data Collection Platform}
\label{fig:label}
\end{figure}
\begin{table}[h]
\centering
\resizebox{\linewidth}{!}{

\begin{tabular}{cccccc}
\hline
Sensor & Modalities & Data Collected \\\hline
1.OS0-128 & Spinning Lidar & Strongest, 2nd Strongest \\
2.Livox Avia & Solid-state Lidar & First, Second, Third \\
3.Hesai QT64 & Spinning Lidar & First, Last\\
4.Insta360 & Fisheye Camera & RGB Image \\\hline
\end{tabular}
}
\caption{Sensors}
\label{tab:sensors}
\end{table}
\subsection{Polar 3D Scanner with Ouster Lidar and Insta360 camera}
The Polar 3D Scanner is equipped with Ouster OS0-128 REV6 Lidar and an Insta360 ONE RS 1-Inch 360 Edition Camera. It also contains an Xsens Mti-630 IMU, enabling it to provide real-time environment scans and previews via mobile phones. Additionally, its post-processing software allows the extraction of colored point clouds and textured meshes, as well as raw RGB and Lidar data.

The Ouster OS0-128 REV6 is an ultra-wide field-of-view spinning Lidar sensor with a 90° vertical field of view and a range of 35m at 10 reflectivity. It has 128 channels of resolution and can output up to 5.2 million points per second. The L2X digital Lidar chip powers all of Ouster’s latest Rev 06 OS series scanning sensors and is capable of counting up to 1 trillion photons per second and outputting up to 5.2 million points per second. The L2X chip can process both the strongest and second strongest returns of light for each point \cite{ouster}.

The Insta360 ONE RS 1-Inch 360 Edition is a dual 1-inch sensor 6K 360 camera that delivers less noise, more detail and better shots. It has dual 1-inch rolling shutter Sony IMX283 sensors that capture 3072x3072 fisheye images for front and rear camera. These large sensors allow the camera to capture more light and detail, resulting in higher quality images and videos \cite{insta360}.

\subsection{Livox Avix Triple Return Lidar}
The Livox Avia is a solid-state LiDAR sensor that combines compact and lightweight design with improved detection range and efficiency. It features an FOV greater than 70° and its triple-echo modes are ideal for use cases such as mapping and low-speed autonomous driving. Livox Avia has a point cloud data rate of up to 240,000 points/s in single return mode \cite{livox}.

\subsection{Pandar Qt64 Dual Return Lidar}
The Hesai Pandar QT64 is a short-range mechanical Lidar equipped with 64 channels of data for ultimate road protection. It has an ultra-wide vertical FOV of 104.2° (-52.1° to +52.1°) and a minimum vertical resolution of 1.45°. We record in dual return mode, which can get the first and last return \cite{hesai}.

\subsection{Time Sync and Calibration}
With the use of the Precision Time Protocol (PTP) for hardware synchronization, all three Lidar timestamps can be hardware synchronized within 1 ms. The Polar scanner components, Ouster OS0-128 Lidar, Xsens IMU, and Insta360 camera, are factory calibrated and synchronized by the manufacturer. An initial extrinsic calibration between three Lidars was provided by registering the all Lidar point cloud to a high-precision ground truth point cloud collected by a FARO scanner in the calibration field. This provides a rough estimate of the Lidar-to-Lidar transforms.

At run-time, the state-of-the-art mulitiple Lidar SLAM algorithm MA-LIO \cite{jung2023asynchronous} performs online extrinsic calibration and temporal synchronization. By modeling the continuous-time trajectory using B-spline interpolation and propagating uncertainties, MA-LIO is able to accurately estimate the time-varying extrinsic calibration between the Lidars. It also compensates for any residual timestamp offsets to synchronize the asynchronous scans.

MA-LIO provides the synchronized trajectory along with undistorted and aligned point clouds from each Lidar in the global frame. By explicitly handling the spatial and temporal discrepancies between Lidars, We can get precise real-time calibration and synchronization without requiring overlapping fields of view or strict hardware synchronization. The resulting point clouds are ready for direct fusion and mapping without any further alignment steps.

\section{Dataset}
\label{sec:dataset}
We collect the Lidar reflection dataset in diverse indoor environments including office spaces, corridor, and rooms containing challenging reflective surfaces. The dataset contains synchronized Lidar point clouds, RGB images, and ground truth annotations.

\subsection{Data Collection Process}
We handheld the multi-sensor platform and walk through the environments, scanning from various angles and ranges. For each scene, we recorded multiple passes sampling different viewpoints to capture reflections.

\subsection{Data Annotation and Labeling}
\begin{figure}[ht]
\centering
\includegraphics[width=0.43\textwidth]{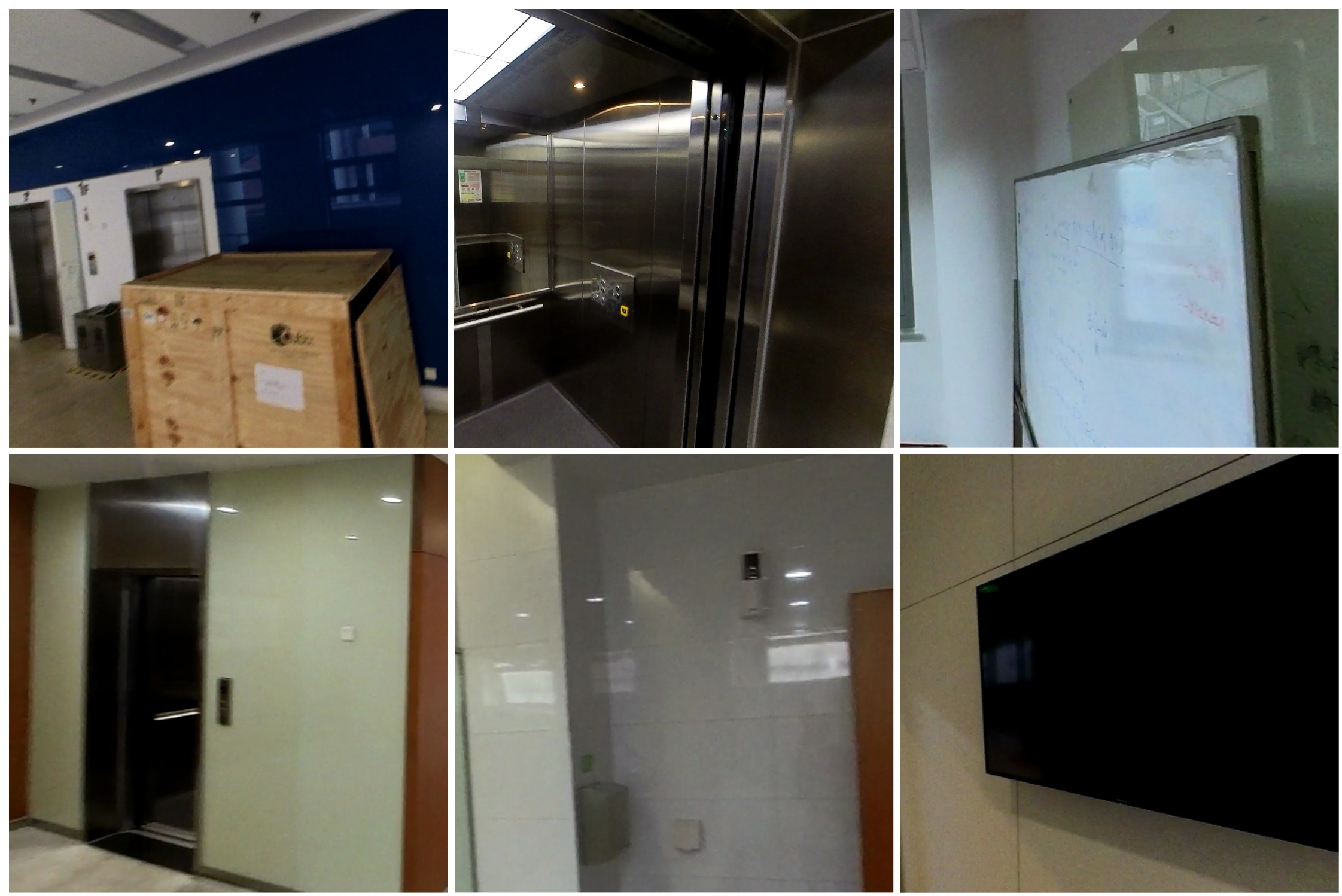}
\caption{Other Reflective Objects}
\label{fig:otherref}
\end{figure}

We propose utilizing a ground truth representation and point cloud labels for the dataset. The ground truth representation encompasses a texture mesh, with ground truth point cloud labels being automatically generated based on the Lidar's pose. The mesh, generated from the Polar scanner, was manually cleaned and fixed holes caused by glass reflections or other issues. Different types of reflections were labeled on the texture images using different colors. We categorized reflective surfaces into three types: glass, mirrors, and other reflective objects. As shown in Figure \ref{fig:otherref}, the last category includes acrylic boards, whiteboards, TV screens, monitors, and glazed tiles. While these objects exhibit a lower reflectivity compared to glass and mirrors, they can still create problematic noise points and holes in Lidar data as well as bright overexposed regions in camera images.

After labeling the ground truth meshes, we utilize Open3D \cite{Zhou2018} to propagate labels to the Lidar point clouds via ray-casting. The pose of each Lidar point cloud is used to transform it into the mesh coordinate frame. Rays are traced from the Lidar origin to each point, and ray-casting is performed against the labeled mesh. Points are then annotated based on the ray-casting results. We assign the following labels:

\begin{table*}[ht]
\centering
\resizebox{\linewidth}{!}{
\begin{tabular}{c|c|cccccc||ccc}
\hline
Data/size & Sensor & Normal & Glass & Mirror & OtherRef & Reflection & Obstacle & 1stReturn  & 2ndReturn  & 3rdReturn  \\ \hline
\multirowcell{3}{Seq 1\\3732} & Ouster & 84.19 & 0.91 & 0.61 & 2.85 & 9.31 & 0.66 & 99.72 & 0.28 & / \\
 & Hesai & 76.67 & 3.16 & 3.41 & 2.60 & 10.77 & 1.54 & 92.60 & 7.40 & / \\
 & Livox & 52.23 & 4.73 & 1.72 & 2.42 & 29.22 & 7.62 & 86.30 & 11.63 & 2.07 \\ \hline
\multirowcell{3}{Seq 2\\4702} & Ouster & 87.75 & 2.59 & 0.04 & 3.23 & 2.41 & 2.45 & 99.71 & 0.29 & / \\ 
 & Hesai & 78.64 & 7.48 & 0.28 & 3.14 & 3.79 & 4.16 & 93.82 & 6.17 & / \\
 & Livox & 68.71 & 6.34 & 0.13 & 1.73 & 11.60 & 9.48 & 92.66 & 6.90 & 0.42 \\ \hline
\multirowcell{3}{Seq 3\\7574} & Ouster & 87.05 & 2.17 & / & 2.76 & 1.67 & 3.64 & 99.62 & 0.38 & / \\
 & Hesai & 76.97 & 7.76 & / & 2.14 & 3.41 & 5.46 & 92.13 & 7.87 & / \\ 
 & Livox & 55.78 & 5.33 & / & 1.29 & 14.34 & 19.28 & 83.49 & 14.33 & 2.17 \\ \hline
\end{tabular}
}
\caption{Percent of Point Label and Return in Each Sequence for each sensor}
\label{tab:dataset}
\end{table*}

The labels categorize the different types of points that can arise in a 3D scan of reflective environments. Label 0 refers to unlabeled points outside the mesh boundary. Label 1 denotes normal non-reflective surface points. The next three labels indicate points on glass, mirror, and other reflective object surfaces, respectively. Label 5 marks reflection points corresponding to virtual objects rather than real surfaces. Finally, label 6 identifies points from surfaces occluded behind transparent objects like glass. To maintain clarity, the color coding of labels in the dataset matches the colors used to visualize labeled points in the figures throughout the following paper.

\begin{center}
\fcolorbox{black}{white}{\rule{0pt}{6pt}\rule{6pt}{0pt}}\textbf{ 0:Unlabeled}
\fcolorbox{black}{gray}{\rule{0pt}{6pt}\rule{6pt}{0pt}}\textbf{ 1:Normal Points}
\fcolorbox{black}{green}{\rule{0pt}{6pt}\rule{6pt}{0pt}}\textbf{ 2:Glass}

\fcolorbox{black}{blue}{\rule{0pt}{6pt}\rule{6pt}{0pt}}\textbf{ 3:Mirrors}
\fcolorbox{black}{cyan}{\rule{0pt}{6pt}\rule{6pt}{0pt}}\textbf{ 4:Other Reflective Objects}

\fcolorbox{black}{magenta}{\rule{0pt}{6pt}\rule{6pt}{0pt}}\textbf{ 5:Reflection Points}
\fcolorbox{black}{yellow}{\rule{0pt}{6pt}\rule{6pt}{0pt}}\textbf{ 6:Obstacle behind Glass}
\end{center}

To generate labels for the RGB images, we render the camera view on the labeled mesh using Open3D, leveraging the camera pose and intrinsic. Masks are extracted based on the rendered texture colors. For compatibility with existing deep learning methods, we undistort the fisheye images to a pinhole camera model using the manufacturer's provided intrinsic.

In total, the dataset contains 48024 labeled point clouds with 16008 point clouds per Lidar sensor across three sequences. Additionally, there are 3799 labeled RGB images across the sequences.

\subsection{Dataset and Statistics}
\begin{figure}[ht]
\centering
\includegraphics[width=0.38\textwidth]{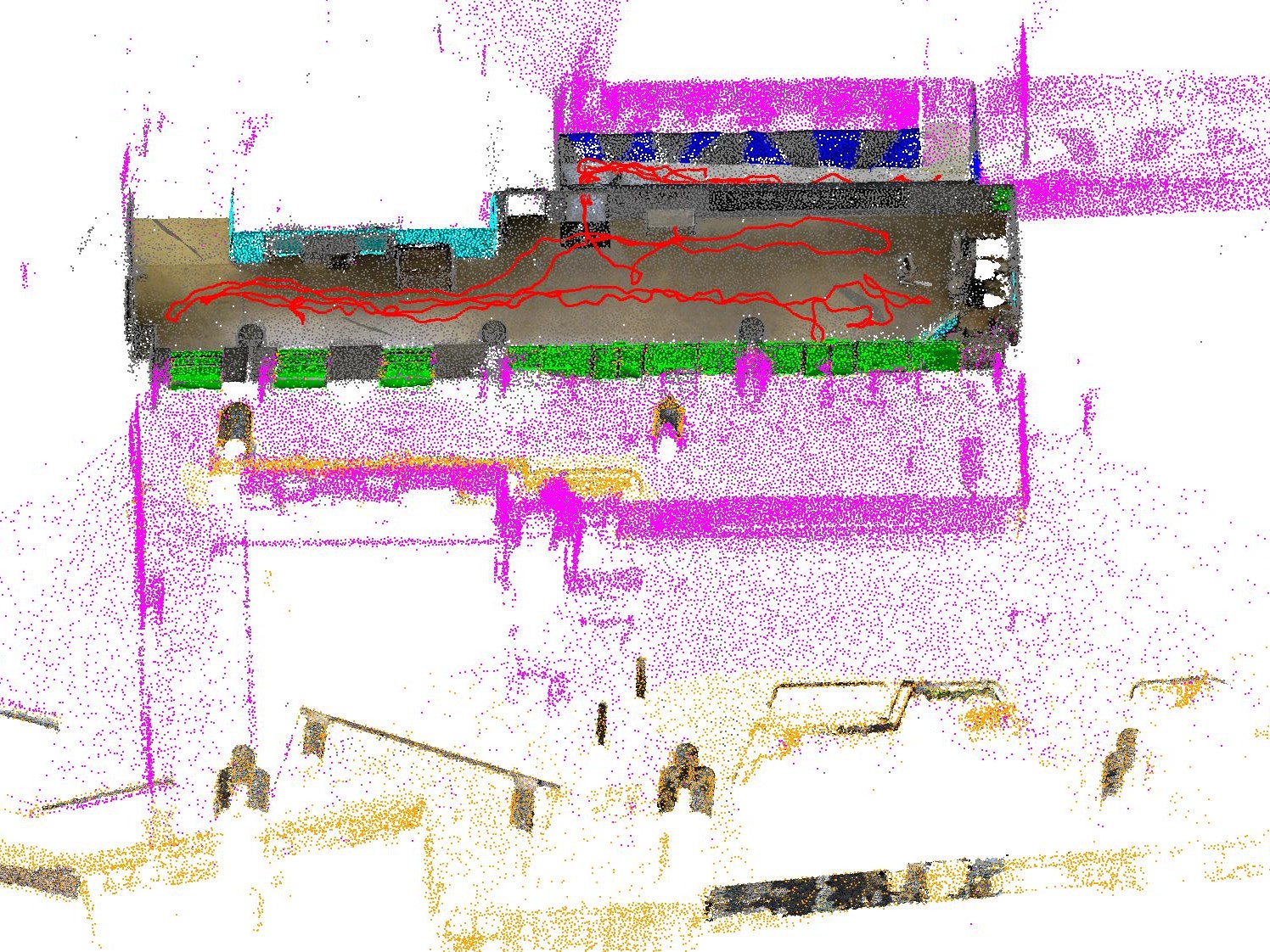}
\caption{Sequence 1: Corridor with Mirrors}
\label{fig:seq1}
\end{figure}

\begin{figure}[ht]
\centering
\includegraphics[width=0.4\textwidth]{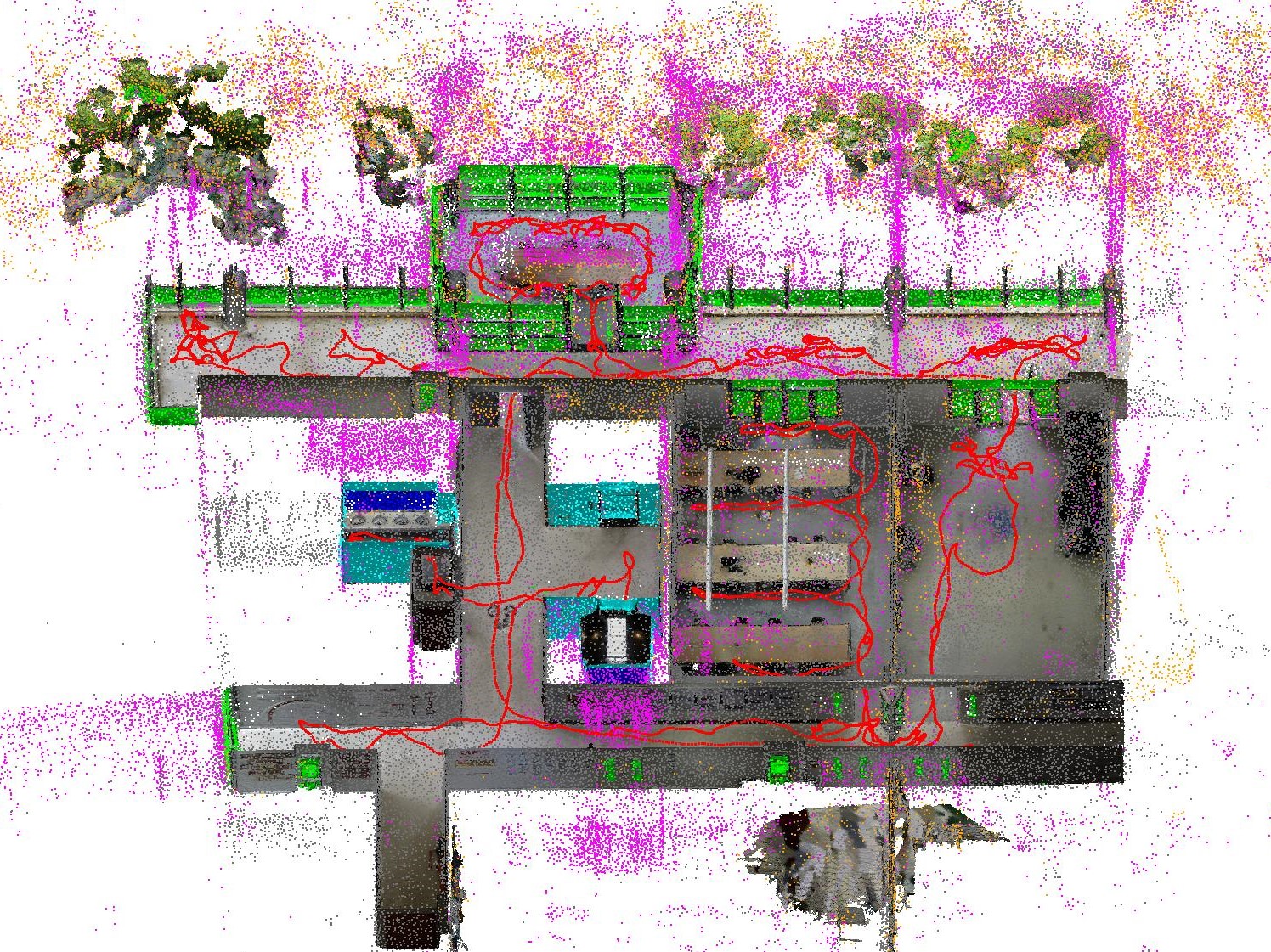}
\caption{Sequence 2: Rooms}
\label{fig:seq2}
\end{figure}

\begin{figure}[ht]
\centering
\includegraphics[width=0.4\textwidth]{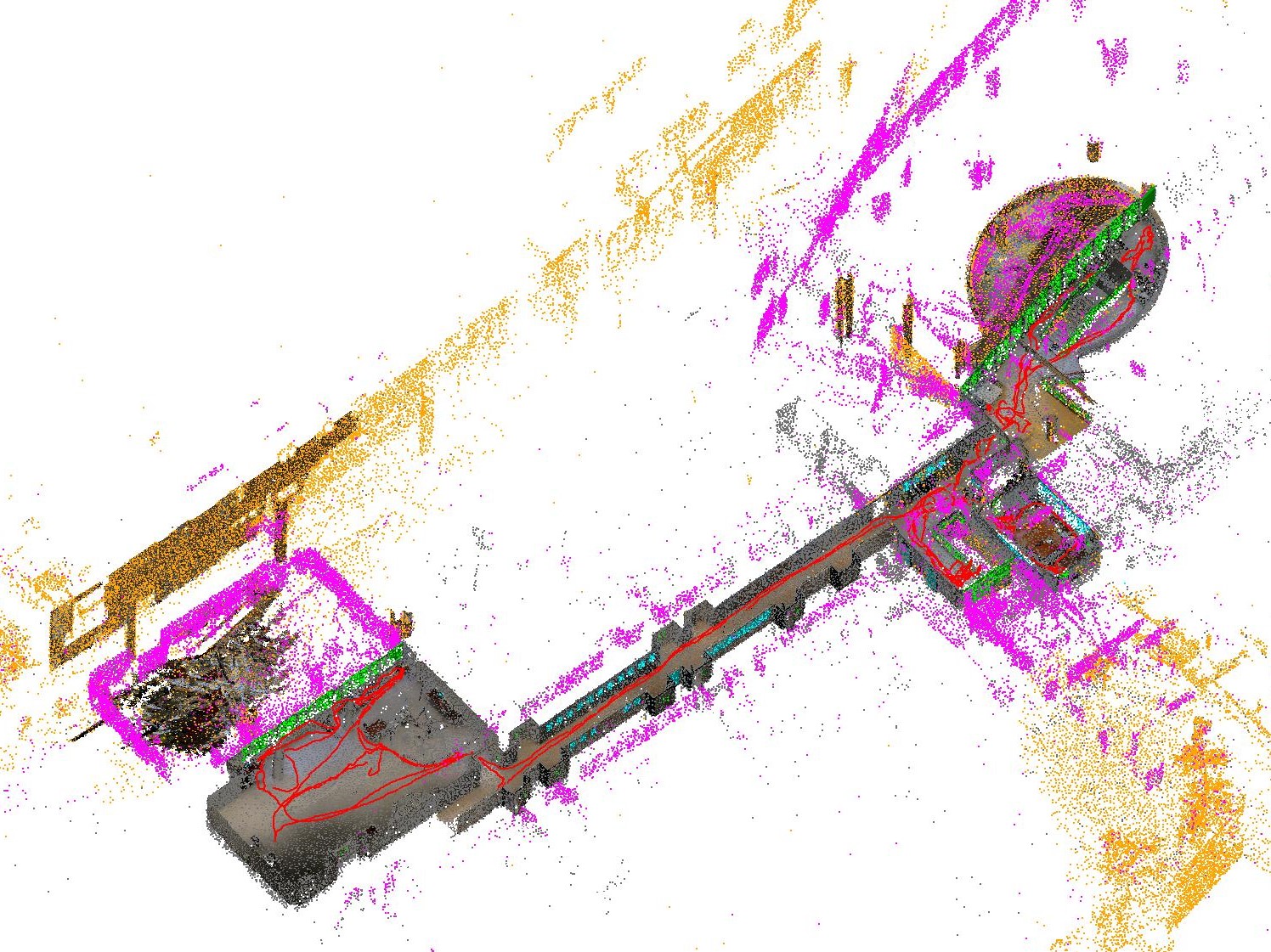}
\caption{Sequence 3: Floor}
\label{fig:seq3}
\end{figure}

The dataset contains three sequences captured in different indoor environments. Table \ref{tab:dataset} provides label statistics and return numbers statistics for each Lidar sensor. We observe that while the Ouster Lidar is dual-return, very few ($<$0.4\%) points have the second return compared to the other Lidars.

We visualize the labeled mesh and point clouds in the Figures \ref{fig:seq1}, \ref{fig:seq2} and \ref{fig:seq3} and describe each sequence below. The images show the defined color coding for each label. The red line indicates the data collection trajectory. For better visualization, point clouds are subsampled to 0.1m resolution and ceiling points are removed.

\subsubsection{Sequence 1: Corridor with Mirrors}
Sequence 1, visualized in Figure \ref{fig:seq1}, contains 3732 point clouds per Lidar and 541 RGB images. It features a corridor with wall-mounted mirrors that reflect the opposite wall and ceiling. Additionally, there is a passage with many floor-to-ceiling windows showing views outside. This sequence provides challenging mirror reflections as well as glass surfaces and other reflective objects.

\subsubsection{Sequence 2: Rooms}

Sequence 2, shown in Figure \ref{fig:seq2}, consists of 4702 point clouds per Lidar and 1716 RGB images captured in an office environment. It includes office rooms, a conference room with glass all-around, hallways with glass doors and railings looking outside, as well as mirrors and other reflective objects. This provides a diverse set of transparent, reflective, and glass surfaces.

\subsubsection{Sequence 3: Floor}
Sequence 3, visualized in Figure \ref{fig:seq3}, comprises 7574 point clouds per Lidar and 1542 RGB images captured on a floor with rooms, meeting rooms, floor-to-ceiling windows, glass railings, and several reflective objects including posters, whiteboards, and wall-mounted TVs. This provides many examples of reflective surfaces beyond just glass and mirrors.

\subsection{Dataset Structure}

The provided dataset includes four main parts: Raw data, SemanticKitti Pointcloud, RGB Image, Scripts. Additionally, we provide trained models on our new dataset to enable out-of-the-box evaluation. The dataset structure is organized as follows:

\begin{itemize}
\item \textbf{Raw}: Contains the raw sensor data for each sequence, including Lidar pose files, images, labeled meshes, origin textured meshes, labeled ray-casting point clouds, camera extrinsic and intrinsic.
\item \textbf{RGB}: Contains the RGB images and masks for different labels (glass, mirror, other reflective, all reflective), split into train and test folders for each. All reflective means labels including glass, mirror and other reflective objects.
\item \textbf{SemanticKitti}: Labeled point clouds in SemanticKitti format with 4 (x,y,z,intensity) and 5 channels (x,y,z,intensity,return) for benchmark.
\item \textbf{Scripts}: Helper scripts for dataset processing tasks like raytracing, statistics and evaluation.
\item \textbf{Network}: Code and weights for reflection detection networks like EBLNet, PCSeg, SATNet.
\end{itemize}

\subsection{Dataset Analysis}
Utilizing the labeled point clouds, we analyzed the multi-return characteristics and the relationship between the laser beam incident angle and different reflection types.
\begin{table}[ht]
\resizebox{\linewidth}{!}{
\begin{tabular}{c|ccc}
\hline
Label      & 1st Return  & 2nd Return  & 3rd Return   \\ \hline
Normal     & 99.02 & 0.97  & 0.01 \\
Glass      & 99.61 & 0.39  & 0.00 \\
Mirror     & 98.74 & 1.26  & 0.00 \\
OtherRef   & 99.82 & 0.18  & 0.00 \\
Reflection & 77.16 & 22.07 & 0.77 \\
Obstacle   & 62.20 & 33.38 & 4.43 \\ \hline
\end{tabular}
}
\caption{The percent of different labels in each return}
\label{tab:return}
\end{table}
The analysis in Table \ref{tab:return} provides insights into detecting reflections using multi-return Lidar. For normal, glass, mirror, and other reflective object points, nearly all points appear in the first return, with under 1\% in later returns. This matches expectations since these kind of surfaces immediately reflect the Lidar beam. The small fraction of later returns may arise from sensor noise.

In contrast, reflection points and obstacles behind glass exhibit substantially higher ratios (22-33\%) in the second return compared to other classes. In the third return, obstacles behind glass dominate, with some remaining reflections. Leveraging the return channel enables identifying more reflections and obstacle behind versus using just the first return. This demonstrates the value of multi-return Lidar data for robust reflection analysis.

\begin{figure}[ht]
\centering
\includegraphics[width=0.45\textwidth]{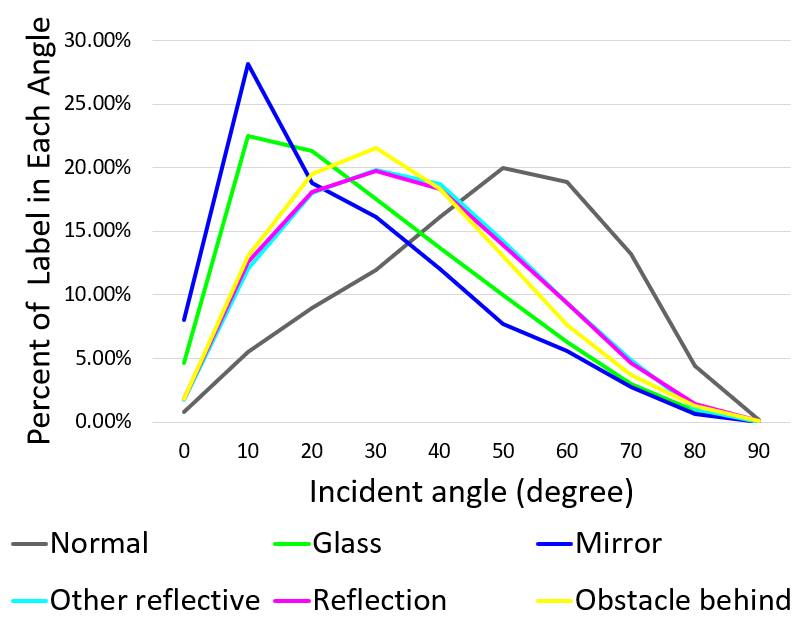}
\caption{Percentage of labels relative to the laser beam incident angle.}
\label{fig:labeldegree}
\end{figure}

\begin{table*}[ht]
\centering
\resizebox{\linewidth}{!}{

\begin{tabular}{c|cc|ccccccc}
\hline
Methods & Return & Sensor & Total & Normal & Glass & Mirror & Other Reflective & Reflection & Obstacle \\ \hline
Minkowski\cite{choy20194d} & N & All & 81.5479 & 96.5204 & 69.395  & 76.5126 & \textbf{84.9613} & 84.4523 & 77.446  \\
SPVCNN\cite{tang2020searching}  & N & All & 81.8484 & 96.5277 & 69.7724 & 77.7727 & 84.1454 & 84.7161 & 78.156  \\
Cylinder3D\cite{zhu2021cylindrical} & N & All & 83.7196 & 96.8398 & 69.7819 & \textbf{79.8688} & 83.7576 & 88.5802 & 83.4894 \\
Minkowsk\cite{choy20194d} & Y & All & 82.1131 & 96.7489 & \textbf{70.2697} & 74.3789 & 84.9468 & 86.4367 & 79.8977 \\
SPVCNN\cite{tang2020searching} & Y & All & 82.1435 & 96.6711 & 69.4181 & 77.6020 & 86.4166 & 84.8909 & 77.8623 \\ 
Cylinder3D\cite{zhu2021cylindrical} &  Y & All & \textbf{83.9188} & \textbf{96.9597} & 70.0164 & 79.6803 & 83.5822 & \textbf{89.2830}  & \textbf{83.9912} \\ \hline

Cylinder3D\cite{zhu2021cylindrical} &  Y & Ouster & 79.7141 & \textbf{97.8982} & 58.2692 & 59.3531 & 88.0816 & \textbf{90.505} & 84.1773 \\
Cylinder3D\cite{zhu2021cylindrical} &  Y & Hesai & 84.9809 & 96.5099 & \textbf{78.6898} & \textbf{88.4585} & 80.8066 & 86.5446 & 78.8758 \\
Cylinder3D\cite{zhu2021cylindrical} &  Y & Livox & \textbf{85.9603} & 93.5877 & 72.6401 & 85.7264 & \textbf{87.8459} & 89.3691 & \textbf{86.5924}
 \\
\hline
\end{tabular}
}
\caption{Lidar Benchmark Results (mIOU)}
\label{tab:results}
\end{table*}

Using the mesh normals and laser beam directions, we compute the incident angle for each point. Figure \ref{fig:labeldegree} plots the percentage of points belonging to each reflective label versus incident angle, with each line add up to 100\% across angles. For normal points, the percentage first increases towards grazing angles then decrease at higher incident angle, as most Lidar beams from wide FOV Lidar do not strike surfaces head-on. In contrast, glass and mirrors have a higher percentage of direct returns occur at lower incident angle where the beam is closer to  perpendicular. As the incident angle increases, fewer specular returns persist. For obstacle behind, other reflective objects and reflection points exhibit higher percentages at moderate incident angle and fewer points at higher angles. This analysis shows that a more frontal Lidar view yields more direct returns from glass and mirrors, while a grazing angle causes more reflections. 


\section{Benchmark Evaluation}
\label{sec:benchmark}

We benchmark various reflection detection methods on our dataset, including Lidar-based and RGB-based approaches. The Lidar-based approach uses point cloud geometry to detect reflective points and categorize them, whereas the RGB-only method relies purely on RGB data for reflection point detection. 

For point cloud analysis, we leverage the open-source PCSeg codebase \cite{pcseg2023}, which implements MinkowskiNet~\cite{choy20194d}, Cylinder3D \cite{zhu2021cylindrical}, and SPVCNN \cite{tang2020searching} segmentation methods. These geometrically analyze the 3D points to detect reflective surfaces.

For RGB analysis, we evaluate the state-of-the-art GlassSemNet \cite{lin2022exploiting} and EBLNet \cite{he2021enhanced} for glass, along with HetNet \cite{he2023efficient} and SATnet \cite{huang2023symmetry} for mirror detection. Some methods provide only pretrained models for testing but without training code, so we directly evaluate these pretrained models on our dataset. We also retrain SATnet and EBLNet on our new reflection-labeled RGB images of glass and all reflective to compare performance gains.

\subsection{Benchmark Setup}
Our benchmarking process is conducted on a computer equipped with dual NVIDIA RTX3090 GPUs, with 24GB of GPU memory each.
For Lidar-based detection, the original SemanticKitti dataset format only provides 4 channels (x, y, z, intensity) per point. To compare the result of adding return information for each point, we made modifications to the PCSeg code to give 5 channels (x, y, z, intensity, return)  per point. For RGB-based detection, we compare the results using a pretrained model and, subsequently, retraining the model using our dataset of glass and alllabels.

\subsection{Benchmark Results and Ablation Study}

\begin{table}[h]
\centering
\begin{tabular}{cccc}
\hline
Methods & Dataset & Model & mIOU \\\hline
GlassSemNet\cite{lin2022exploiting} & GDD &  V2 & 90.80 \\
GlassSemNet\cite{lin2022exploiting} & 3DRef-Glass &  V2 & 53.69 \\
HetNet\cite{he2023efficient} & PMD &  PMD& 69.00 \\
HetNet\cite{he2023efficient} & 3DRef-Mirror &  PMD& 44.05 \\
SATNet\cite{huang2023symmetry} & RGBD &  RGBD & 78.42 \\
SATNet\cite{huang2023symmetry} & 3DRef-Mirror &  RGBD & 49.46 \\
EBLNet\cite{he2021enhanced} & GDD &  GDD & 88.72 \\
EBLNet\cite{he2021enhanced} & 3DRef-Glass &  GDD & 60.49 \\
EBLNet\cite{he2021enhanced} & MSD &  MSD & 80.33 \\ 
EBLNet\cite{he2021enhanced} & 3DRef-Mirror &  MSD & 57.61 \\ \hline
*SATNet\cite{huang2023symmetry} & 3DRef-Mirror &  / & \textbf{82.47} \\
*SATNet\cite{huang2023symmetry} & 3DRef-All &  / & 68.81 \\
*EBLNet\cite{he2021enhanced} & 3DRef-Glass &  / & \textbf{86.71} \\
*EBLNet\cite{he2021enhanced} & 3DRef-All &  / & \textbf{87.60} \\\hline

\end{tabular}
\caption{RGB Benchmark Results. Model refers to the pretrained model name. Asterisk (*) denotes networks retrained on the 3DRef dataset.}
\label{tab:resultsrgb}
\end{table}

Table \ref{tab:results} shows the benchmark results for Lidar-based reflection detection methods using just XYZI (4 channels) versus adding the return channel (XYZIR, 5 channels).

Adding the explicit return channel improves mIOU by ~0.5\% across methods, demonstrating its value for identifying reflective points. However, the gain is limited since some Lidars (e.g. Ouster) have very few dual returns and different Lidar has different retrun type. Still, leveraging multi-return patterns enables detecting more reflections.

Among sensors, Livox achieves the highest accuracy, while Ouster struggles on glass and mirror classes. As shown in Table \ref{tab:dataset}, Ouster initially detects fewer glass and mirror points compared to the other Lidars. This likely contributes to its lower detection rate on those classes. Overall these results demonstrate multi-return Lidar's advantages for analyzing reflective surfaces.

Table \ref{tab:resultsrgb} benchmarks RGB methods using default pretrained models versus retraining on 3DRef.
Retraining substantially improves performance for glass detection. EBLNet's glass mIoU increases from 60.49\% to 86.71\% after retraining. For mirrors, retrained SATNet's mIoU rises from 49.46\% to 82.47\%. This highlights the domain gap between existing datasets and our new benchmark. Retrained SATNet also achieves a strong 87.6\% mIoU on all reflective classes, indicating it generalizes well to diverse reflections.

In summary, retraining on our large-scale multi-modal dataset leads to major performance boosts, confirming its value for advancing reflection detection networks. The variety of reflective environments and aligned multi-sensor data enables robust models that can handle real-world deployment challenges.

\section{Conclusion}
\label{sec:conclusion}

This work introduces a large-scale multi-modal 3D dataset to advance robust reflection detection, enabling reliable perception to reflective surfaces. The diverse environments, precise ground truth annotations, and analysis of current methods confirm the value of leveraging aligned Lidar and RGB data. Detailed benchmarks assess the performance of current Lidar point cloud and RGB image segmentation methods, providing insights into factors like multi-return analysis. Detailed benchmarks demonstrate significant performance gains from retraining models on this data compared to pretrained networks. This highlights the need for comprehensive reflective data to handle real-world deployment challenges.

Several fruitful directions exist for future work. Expanding the diversity of datasets across sensors, materials, and environments will further boost detection robustness. Additionally, fusing Lidar and RGB reflection detection networks is a promising approach to combine geometric and semantic cues. Exploring different sensor modalities like depth and polarization represents another opportunity.  Moving beyond supervised learning with self-supervised and semi-supervised techniques is also valuable.

This benchmark dataset lays the groundwork to drive future research towards reflection disambiguation and robust 3D mapping. The comprehensive testbed will continue driving future research and methods towards reflection disambiguation. Enabling autonomous robots and vehicles to accurately perceive and map reflective environments will be key to unlocking real-world operation.

\textbf{Acknowledgements:} This work has been partially funded by the Shanghai Frontiers Science Center of Human-centered Artificial Intelligence. This work was also supported by the Science and Technology Commission of Shanghai Municipality (STCSM), project 22JC1410700 ”Evaluation of real-time localization and mapping algorithms for intelligent robots”.

{
    \small
    \bibliographystyle{ieeenat_fullname}
    \bibliography{main}
}

\end{document}